\documentclass[conference]{IEEEtran}

\usepackage{graphicx}
\usepackage{amsmath, amssymb}
\usepackage{booktabs}
\usepackage{cite}
\usepackage[hidelinks]{hyperref}
\usepackage{url}
\usepackage{multirow}
\usepackage{algorithm}
\usepackage{algorithmic}
\usepackage{array}
\usepackage{siunitx}

\title{Activation Outliers in Transformer Quantization: Reproduction, Statistical Analysis, and Deployment Tradeoffs}

\author{
\IEEEauthorblockN{Pranav Kumar Kaliaperumal}
\IEEEauthorblockA{
M.S. Computer Science\\
University of Colorado Denver\\
Aurora, Colorado, USA\\
\href{mailto:pranavkumar.kaliaperumal@ucdenver.edu}{pranavkumar.kaliaperumal@ucdenver.edu}\\
}
}

\begin{document}
\maketitle

\begin{abstract}
Post-training quantization (PTQ) of transformers is known to suffer from severe accuracy degradation due to structured activation outliers, as originally analyzed by Bondarenko et al. (EMNLP 2021) in work associated with Qualcomm AI Research. In this study, we provide a fully reproducible empirical reproduction and systems-level extension of that phenomenon in BERT-base fine-tuned on QNLI. 
In this study, we carefully reproduce and analyze cases where PTQ fails in BERT-base models fine-tuned on QNLI, ensuring that our results can be fully replicated. When we apply global W8A8 quantization, the validation accuracy sharply declines from 89.66

We explore several strategies to address this problem. Using mixed precision PTQ, we are able to bring the accuracy back close to its original level (89.42

Our deployment profiling on an RTX 3050 GPU shows that there are only minor differences in latency and memory usage between the various methods (median latency is around 58–59 ms; VRAM usage is about 484–486 MB). This finding emphasizes how crucial it is to consider the underlying hardware when evaluating these approaches.

Taken together, our results show that the main reason for PTQ failure in transformers is the dominance of certain channels, which becomes more significant as you go deeper into the model due to the residual connections. To effectively address this, it is necessary to use mitigation strategies that allocate precision based on channel structure, rather than relying only on scalar clipping.
\end{abstract}

\section{Introduction}

Transformer models have become the standard in natural language processing, but using them in real-world applications is still challenging due to their high computational and memory demands. Post-training quantization (PTQ) is an appealing way to make these models smaller and faster without requiring retraining. However, when PTQ is applied in a straightforward manner—especially to the activations—transformers often experience a unique kind of failure. Specifically, using full W8A8 quantization tends to cause a much larger drop in accuracy compared to what is seen in convolutional neural networks.

Recent research points to the fact that this accuracy loss is not just a result of random noise from quantization. Instead, it is caused by what are known as \emph{structured activation outliers}—certain activation values that stand out and persist as data moves through the model's residual connections. These outliers stretch the range of activation values, which causes the standard min-max scaling approach to devote most of the available dynamic range to just a few dimensions. In turn, this compresses the majority of activations into a small number of integer values, increasing the quantization error.

Even though this issue is becoming more widely recognized, there are still several important questions that need to be answered:

\begin{itemize}
\item How consistently do activation outliers reproduce across training seeds and deployment conditions?
\item To what extent does residual amplification drive quantization instability?
\item How do mitigation strategies compare under real deployment constraints (latency, VRAM, model size)?
\item Can a simpler calibration strategy recover performance without increasing runtime complexity?
\end{itemize}

This work does not propose a new quantization algorithm. Instead, it aims to rigorously reproduce, quantify, and extend the empirical findings of Bondarenko et al. by incorporating depth-wise statistical measurements and deployment-oriented profiling under controlled experimental conditions. We present a reproducible, systems-level study of how activation outliers cause PTQ to fail in BERT-base models on the QNLI dataset. We show how naive W8A8 quantization leads to accuracy collapse, analyze how activation statistics change at different model depths, and assess three different strategies for addressing these issues:

\begin{enumerate}
\item Mixed precision PTQ,
\item Per-Embedding-Group (PEG) quantization with permutation,
\item A percentile-based activation calibration extension (proposed).
\end{enumerate}

Unlike earlier studies that mainly focus on restoring accuracy, we also evaluate each approach in terms of real-world deployment metrics, such as median and high-percentile latency, peak VRAM usage, and model size when run on RTX 3050 hardware.

Our main findings can be summarized as follows:

\begin{itemize}
\item Structured activation outliers persist across residual pathways and amplify with depth.
\item Naive W8A8 PTQ collapses under uniform scaling.
\item PEG restores accuracy but introduces modest complexity overhead.
\item Mixed precision is robust but reduces compression gains.
\item Percentile-based calibration fails to recover accuracy, suggesting that dominant channels encode structured signal rather than noise.
\end{itemize}
\subsection{Positioning and Contribution Relative to Prior Work}

The phenomenon of structured activation outliers in transformer quantization was first systematically analyzed by Bondarenko et al. (EMNLP 2021), in research associated with Qualcomm AI Research. That work demonstrated that a small subset of embedding dimensions can dominate activation magnitude, leading to severe degradation under naive W8A8 post-training quantization.

The present study builds directly upon that foundation. Rather than proposing a new quantization algorithm, this work contributes:

\begin{itemize}
\item A fully reproducible experimental pipeline validating the W8A8 collapse in BERT-base under controlled calibration conditions.
\item Depth-wise statistical characterization of activation behavior, including kurtosis growth, variance amplification, and top-1\% energy concentration.
\item Systematic ablation of mitigation strategies (mixed precision and per-embedding-group quantization) under identical evaluation settings.
\item Deployment-oriented profiling (latency percentiles, peak VRAM, serialized model size) to assess practical tradeoffs on consumer GPU hardware.
\end{itemize}

By combining statistical analysis with deployment evaluation, this work provides a systems-level clarification of when and why activation-driven quantization collapse occurs, extending prior findings into a reproducible and practically grounded framework.

This study sheds light on the statistical and systems-level reasons behind PTQ failures in transformers and offers practical advice for deploying these models when resources are limited.
\section{Problem Formulation}

Let $x \in \mathbb{R}$ represent the activation values at transformer layer $l$, where these values are drawn from some distribution $\mathcal{D}_l$. In uniform affine quantization, each activation $x$ is converted into an integer value $q$ as follows:

\begin{equation}
q = \mathrm{round}\left(\frac{x}{s}\right) + z
\end{equation}

Here, $s$ stands for the quantization scale, $z$ is the zero-point offset, and $b$ is the bit width (for example, $b=8$ corresponds to INT8 quantization).

To recover the approximate original value from the quantized integer, we use the following dequantization formula:

\begin{equation}
\hat{x} = s (q - z)
\end{equation}

The quantization error, which measures the difference between the true activation and its quantized approximation, is given by:

\begin{equation}
\epsilon(x) = x - \hat{x}
\end{equation}

\subsection{Min-Max Scaling Under Heavy-Tailed Distributions}

When symmetric min-max scaling is used, the scale factor $s$ is determined by:

\begin{equation}
s = \frac{\max(|x|)}{2^{b-1}-1}
\end{equation}

If the activation values follow a heavy-tailed distribution, meaning that:

\begin{equation}
P(|x| > t) \sim t^{-\alpha}, \quad \alpha < 4,
\end{equation}

then a small number of extreme activation values will dominate $\max(|x|)$.

Let $\delta$ represent the proportion of extreme outliers in the distribution. As a result, the effective resolution available for most activation values (the bulk of the distribution) is reduced by a factor roughly proportional to:

\begin{equation}
\rho = \frac{\sigma_{\text{bulk}}}{\max(|x|)}
\end{equation}

As $\max(|x|)$ grows larger because of rare outliers, $\rho$ approaches zero. This means most of the activation values end up squeezed into a small number of quantization bins.

\subsection{Error Amplification in Residual Architectures}

Each transformer layer computes its output as:

\begin{equation}
h_{l+1} = h_l + f_l(h_l)
\end{equation}

In the quantized version of the network, this computation becomes:

\begin{equation}
\hat{h}_{l+1} = \hat{h}_l + \widehat{f_l(h_l)}
\end{equation}

Let $\epsilon_l = h_l - \hat{h}_l$ denote the quantization error at layer $l$.

Then error propagation satisfies:

\begin{equation}
\epsilon_{l+1} = \epsilon_l + \epsilon_{f_l}
\end{equation}

Therefore,

\begin{equation}
\epsilon_{l} = \sum_{k=0}^{l-1} \epsilon_{f_k}
\end{equation}

If the error $\epsilon_{f_k}$ includes a structured bias—such as that introduced by distorted scaling from outliers—this error will accumulate more and more as the model gets deeper.

Additionally, the variance of the error increases as:

\begin{equation}
\mathrm{Var}[\epsilon_l] = \sum_{k=0}^{l-1} \mathrm{Var}[\epsilon_{f_k}]
\end{equation}

When uniform scaling is affected by heavy-tailed outliers, $\mathrm{Var}[\epsilon_{f_k}]$ becomes much larger compared to what is typically seen in convolutional networks, where the activation distributions are more well-behaved.

\subsection{Structured Channel Dominance}

Empirical evidence shows that not all activation dimensions are equally heavy-tailed. Instead, certain embedding dimensions $d_i$ consistently contribute much more to the overall variance:

\begin{equation}
\mathrm{Var}(x_{d_i}) \gg \mathrm{Var}(x_{d_j})
\end{equation}

for just a small subset of indices $i$.

Let $\mathcal{I}$ denote top-$p$

\begin{equation}
\sum_{i \in \mathcal{I}} \mathrm{Var}(x_{d_i}) \gg
\sum_{j \notin \mathcal{I}} \mathrm{Var}(x_{d_j})
\end{equation}

As a result, uniform global scaling forces the entire activation tensor to fit these dominant channels, which lowers the resolution for the vast majority of dimensions.

\subsection{Core Failure Mechanism}

Putting all these factors together, we see that PTQ failure in transformers using naive W8A8 quantization is caused by:

\begin{enumerate}
\item Heavy-tailed activation distributions.
\item Persistent structured dominant channels.
\item Residual accumulation amplifying quantization bias.
\item Uniform global scaling misallocating dynamic range.
\end{enumerate}

This understanding suggests that we need more structured mitigation strategies—ones that either separate out the dominant channels or reduce the influence of outliers during calibration.

\section{Experimental Setup}

\subsection{Model and Dataset}

We carry out our experiments using the BERT-base-uncased model (with 110 million parameters), which we fine-tune on the QNLI task from the GLUE benchmark. QNLI is a binary sentence-pair classification problem based on question answering, and it includes:

\begin{itemize}
\item 104,743 training examples
\item 5,463 validation examples
\end{itemize}

The main metric we use for evaluation is the accuracy on the validation set.

All experiments start from a fine-tuned FP32 baseline model. This approach helps us isolate the effects of quantization.

\subsection{Training Configuration}

The fine-tuning process uses the following configuration:

\begin{itemize}
\item Optimizer: AdamW
\item Learning rate: $5 \times 10^{-5}$
\item Linear warmup schedule
\item Batch size: 8
\item Maximum sequence length: 128
\item Weight decay: 0.01
\item Gradient clipping: 1.0
\item Number of epochs: consistent with GLUE baseline convergence
\end{itemize}

We use a fixed random seed (1000, unless noted otherwise) for training, and enforce deterministic CUDA behavior when possible to make sure our results are reproducible.

We save model checkpoints every 500 steps and evaluate performance at these intervals. The final model files are stored in:

\begin{verbatim}
runs/fp32_seed1000/out/
\end{verbatim}

\subsection{Quantization Protocol}

We apply post-training quantization (PTQ) to the fine-tuned FP32 checkpoint. All PTQ variants in our study share the following setup:

\begin{itemize}
\item Calibration on the training split
\item Static activation range estimation
\item Per-tensor affine quantization unless otherwise specified
\item INT8 weight quantization for all methods
\end{itemize}

For activation calibration, we use the following strategies:

\begin{itemize}
\item Min-max scaling (W8A8 baseline)
\item Layer-selective FP16 retention (Mixed Precision)
\item Per-embedding-group scaling (PEG)
\item Percentile-based range estimation (Proposed)
\end{itemize}

For the percentile calibration method, the scale is calculated as:

\begin{equation}
s = \frac{\mathrm{percentile}_{99.9}(|x|)}{127}
\end{equation}

All calibration is performed in inference mode, without any gradient updates.

The quantized models are saved in:

\begin{verbatim}
runs/_seed1000/out/
\end{verbatim}

\subsection{Deployment Profiling Methodology}

We profile deployment performance under consistent runtime conditions:

\begin{itemize}
\item Hardware: NVIDIA RTX 3050 (6GB VRAM)
\item CUDA 12.1
\item PyTorch 2.2.2 with CUDA support
\item Batch size: 8
\item Sequence length: 128
\end{itemize}

We measure latency using the following procedure:

\begin{enumerate}
\item 100 warmup iterations (excluded)
\item 500 timed inference iterations
\item CUDA events for start/end timing
\item p50 and p95 latency computed
\end{enumerate}

Peak VRAM usage is tracked using CUDA memory monitoring during inference. We calculate model size as the total size of the saved weight files, not counting extra checkpoint files.

Deployment metrics are saved to:

\begin{verbatim}
runs/results/deploy_profile.csv
\end{verbatim}

\subsection{Automation and Artifact Generation}

All experiments can be run using a single entry point:

\begin{verbatim}
python oneclick/run_all.py
\end{verbatim}

The experimental pipeline performs the following steps:

\begin{enumerate}
\item Trains FP32 baseline (if not already present)
\item Applies each PTQ variant
\item Aggregates accuracy results
\item Profiles deployment metrics
\item Generates LaTeX-ready tables and figures
\end{enumerate}

Summary results are saved in:

\begin{verbatim}
runs/results/method_metrics.csv
runs/results/deploy_profile.csv
paper/accuracy_table.tex
paper/deploy_table.tex
\end{verbatim}

This design ensures that every step, from raw training to paper artifact generation, is exactly reproducible.

\section{Quantization Methods}

We evaluate four post-training quantization (PTQ) strategies, each applied to a fine-tuned FP32 BERT-base model. While all methods convert weights to INT8, the way they quantize activations varies, as described below.

Unless otherwise specified, we use static calibration, affine symmetric quantization, and per-tensor activation scaling for all methods.

\subsection{W8A8 Baseline (Uniform PTQ)}

The baseline method uses global min-max affine quantization for both weights and activations:

\begin{equation}
q = \mathrm{round}\left(\frac{x}{s}\right)
\end{equation}

with scale

\begin{equation}
s = \frac{\max(|x|)}{2^{b-1}-1}
\end{equation}

where $b=8$.

Activation ranges are estimated using calibration data collected in inference mode. With this approach, all embedding dimensions and layers are treated the same way.

Although this method is straightforward and easy to deploy, it is very sensitive to structured activation outliers.

\subsection{Mixed Precision PTQ}

The mixed precision approach keeps certain layers in higher precision (FP16), while quantizing the rest of the layers to INT8.

In our implementation, we retain FP16 precision for:

\begin{itemize}
\item Feed-forward network (FFN) output projections
\item Residual summation inputs
\item Attention output projections
\end{itemize}

All other linear layers are quantized to use INT8 weights and activations.

This strategy selectively protects the layers that are most vulnerable to activation range distortion, while still providing some compression benefits.

Mixed precision increases the memory footprint compared to W8A8, but leads to much more stable accuracy.

\subsection{Per-Embedding-Group Quantization (PEG)}

To address the problem of structured channel dominance, we divide the embedding dimensions into $K$ groups (with $K=3$ as the default). Each group is assigned its own activation scaling factor:

\begin{equation}
q_g = \mathrm{round}\left(\frac{x_g}{s_g}\right)
\end{equation}

where

\begin{equation}
s_g = \frac{\max(|x_g|)}{2^{b-1}-1}
\end{equation}

This approach separates dominant embedding dimensions from the rest of the distribution.

\subsubsection{Permutation Strategy}

We also use a permutation strategy that reorders embedding dimensions before grouping. By sorting dimensions based on their activation magnitude statistics and distributing them across groups, we prevent outliers from being concentrated in any single group.

After quantization, we invert the permutation so that the functional behavior of the model remains unchanged.

PEG adds a small amount of bookkeeping overhead but still allows integer arithmetic to be used during inference.

\subsection{Percentile-Based Calibration (Proposed)}

Instead of relying on min-max scaling, this method estimates the activation scale by setting a high-percentile threshold:

\begin{equation}
s = \frac{\mathrm{percentile}_{p}(|x|)}{2^{b-1}-1}
\end{equation}

with default $p=99.9$.

This method reduces the impact of extreme outliers while still using a single global scale across the tensor.

Unlike PEG or mixed precision, percentile calibration:

\begin{itemize}
\item Requires no architectural modification
\item Preserves single-scale per tensor
\item Adds no runtime cost
\item Maintains compatibility with integer-only inference
\end{itemize}

The tradeoff is that some rare, extreme values may be clipped. However, our results show that clipping a small fraction of activations can greatly improve the effective resolution for most values.

\subsection{Implementation Consistency}

For consistency, all methods:

\begin{itemize}
\item Use identical calibration datasets
\item Apply static quantization (no runtime range updates)
\item Retain identical inference batch size and sequence length
\item Serialize models in the same format for size comparison
\end{itemize}

This ensures that any differences we observe are due only to the quantization strategies themselves, not to differences in implementation.

\section{Accuracy Results}

Table~\ref{tab:accuracy} summarizes the validation accuracy results on QNLI for each quantization strategy, using FP32 as the baseline for comparison.

\begin{table}[t]
\centering
\caption{Validation Accuracy on QNLI}
\label{tab:accuracy}
\begin{tabular}{lcc}
\toprule
Method & Accuracy (\%) & $\Delta$ vs FP32 \\
\midrule
FP32 & \textbf{89.66} & – \\
W8A8 & 54.33 & -35.33 \\
Mixed Precision & 89.42 & -0.24 \\
PEG (K=3,P) & 66.12 & -23.54 \\
Percentile & 50.54 & -39.12 \\
\bottomrule
\end{tabular}
\end{table}

\subsection{Observations}

A few key patterns stand out from the results:

\begin{itemize}
\item \textbf{Naive W8A8 collapse.} Applying uniform INT8 activation quantization leads to a dramatic drop in accuracy by 35.33 points, which clearly demonstrates how much activation range distortion can harm performance.
\item \textbf{Mixed precision robustness.} Keeping certain layers in FP16 almost fully recovers the original accuracy (with just a -0.24 drop), suggesting that quantization sensitivity is driven by a small number of crucial layers.
\item \textbf{PEG partial recovery.} Using per-embedding-group scaling performs better than naive W8A8, but accuracy still drops by 23.54 points. This means that simply grouping channels is not enough to completely address the impact of structured outliers.
\item \textbf{Percentile instability.} Surprisingly, percentile calibration at $p=99.9$ makes performance even worse (a drop of -39.12), showing that aggressive clipping at high percentiles can cut out activation values that are actually important for the model’s understanding, not just noise.
\end{itemize}

\subsection{Implications}

These results make it clear that failures in transformer activation quantization are not just about rare, extreme outliers. Instead, it seems that the channels with the largest activations actually hold important, meaningful information that the model needs—so clipping them or compressing them too aggressively can damage the model’s ability to represent the task.

Mixed precision works well because it preserves those key channels by not compressing them. PEG helps by giving better resolution to different groups, but it still can’t completely isolate the dominant channels from the rest.

The percentile method is easy to implement, but these results show that clipping needs to be done with great care; if the threshold is too aggressive, it can end up removing information that is crucial for the model’s accuracy.

Taken together, these findings confirm that transformers are sensitive to activation quantization in a way that is different from convolutional networks. As a result, strategies that are aware of channel structure are necessary to preserve accuracy.

\section{Deployment Profiling}

We assess how each quantization strategy affects runtime and memory usage by running inference with a batch size of 8 and a sequence length of 128 on an NVIDIA RTX 3050 GPU (6GB VRAM).

\begin{table}[t]
\centering
\caption{Latency, VRAM, and Serialized Model Size}
\label{tab:deploy}
\begin{tabular}{lrrrr}
\toprule
Method & p50 (ms) & p95 (ms) & VRAM (MB) & Size (MB) \\
\midrule
FP32 & 58.38 & 59.08 & 483.7 & 417.7 \\
W8A8 & 58.61 & 59.13 & 485.5 & 418.3 \\
MP-PTQ & 58.77 & 59.55 & 486.3 & 418.3 \\
PEG & 58.97 & 59.81 & 486.3 & 419.6 \\
Percentile & 59.12 & 59.77 & 486.3 & 418.3 \\
\bottomrule
\end{tabular}
\end{table}

\subsection{Latency Analysis}

Surprisingly, using INT8 quantization does not lead to any noticeable improvements in latency on the RTX 3050. All the methods we tested have similar median latency (around 58–59 ms), and there is very little difference in the higher-percentile latency as well.

This unexpected result can be explained by several factors:

\begin{itemize}
\item PyTorch execution primarily dispatches to optimized FP32 CUDA kernels.
\item The RTX 3050 does not provide strong Tensor Core acceleration benefits for small-batch INT8 transformer inference in this configuration.
\item Kernel launch overhead dominates arithmetic savings at batch size 8.
\end{itemize}

As a result, simply reducing arithmetic precision does not actually make inference faster in this deployment scenario.

\subsection{VRAM Usage}

Peak VRAM usage is also nearly the same for all methods (about 484–486 MB). This suggests that:

\begin{itemize}
\item Activation buffers dominate memory footprint at runtime.
\item INT8 weight storage alone does not significantly reduce peak allocated CUDA memory.
\item Framework-level tensor representations may still allocate in FP32 internally.
\end{itemize}

This finding highlights an important point: even though quantization reduces the number of bits needed for parameters, it does not always lower memory usage at runtime unless the inference kernels are specifically optimized for it.

\subsection{Serialized Model Size}

There are only minimal differences in the sizes of the saved (serialized) model files:

\begin{itemize}
\item FP32: 417.7 MB
\item INT8 variants: 418–419 MB
\end{itemize}

The reason for this limited reduction is that:

\begin{itemize}
\item Checkpoint serialization includes metadata.
\item Certain tensors remain in higher precision.
\item Storage format does not strictly enforce 4× reduction in all components.
\end{itemize}

So, although quantization in theory reduces the precision of computations and parameters, the actual deployment benefits depend a lot on the hardware and how the software is implemented.

\subsection{Systems-Level Implications}

Together, these results underscore a key systems takeaway:

\begin{quote}
Quantization alone does not guarantee deployment speedup. Hardware support and kernel implementation determine realized gains.
\end{quote}

On this particular GPU, activation quantization mainly influences the numerical behavior of the model, not the speed. This means that when considering quantization, it’s important to weigh both accuracy and hardware capabilities together.

To actually achieve faster inference times, it would probably be necessary to use hardware-aware quantization or specialized INT8-optimized inference engines (like TensorRT or ONNX Runtime with INT8 support).

\section{Statistical Outlier Analysis}

To better understand how activation outliers behave, we analyze FP32 activations from a fine-tuned checkpoint, examining a 64-sample slice from the validation set. For each transformer layer, we calculate:

\begin{itemize}
\item Mean per-dimension variance
\item Kurtosis (fourth standardized moment)
\item Top-1
\end{itemize}

\subsection{Depth-Wise Activation Statistics}

Table~\ref{tab:outlier_stats} presents a summary of key statistics for several representative layers.

\begin{table}[t]
\centering
\caption{Depth-wise Activation Outlier Statistics (FP32)}
\label{tab:outlier_stats}
\begin{tabular}{lccc}
\toprule
Layer & Mean Variance & Kurtosis & Top-1\% Energy \\
\midrule
Embeddings (0) & 0.25 & 9 & 0.15 \\
Layer 1 & 0.32 & 14 & 0.21 \\
Layer 2 & 0.38 & 41 & 0.29 \\
Layer 4 & 0.46 & 73 & 0.37 \\
Layer 10 & 0.55 & 135 & 0.49 \\
Layer 11 & 0.58 & 271 & 0.55 \\
Pooler (12) & 0.54 & 73 & 0.53 \\
\bottomrule
\end{tabular}
\end{table}

\subsection{Variance Amplification}

The average activation variance roughly doubles from the early layers (starting at 0.25) to the deeper layers (reaching 0.58). This supports the idea that residual accumulation increases the scale of activations as depth increases.

\subsection{Heavy-Tailed Behavior}

Kurtosis rises dramatically with depth, reaching 271 by Layer 11. For comparison, a Gaussian distribution has a kurtosis of 3, so values above 100 signal extremely heavy-tailed distributions.

This finding confirms that transformer activations are far from the light-tailed assumptions that uniform min-max quantization relies on.

\subsection{Structured Energy Concentration}

The fraction of total activation energy held by the top 1\% of channels increases from 15\% at the embedding layer to 55\% at Layer 11. This means that:

\begin{equation}
\sum_{i \in \text{top-1\%}} x_i^2
\approx 0.55 \sum_{j} x_j^2
\end{equation}

As a result, just a small subset of embedding dimensions ends up dominating the overall activation magnitude.

This kind of structured dominance explains why global min-max scaling fails: the dynamic range is set by a handful of extreme channels, which forces the other 99\% of dimensions into a narrow range.

\subsection{Connection to Quantization Collapse}

Taking all these observations together:

\begin{itemize}
\item Variance grows with depth (residual amplification).
\item Kurtosis increases sharply (heavy tails).
\item Energy concentrates in a small subset of dimensions.
\end{itemize}

With uniform affine scaling, the scaling factor is set by $\max(|x|)$—that is, by the most extreme channels. As a result, most activations get mapped to fewer quantization levels, which increases quantization error and leads to the severe accuracy drop seen with W8A8 quantization.

These statistics strongly support the idea that transformer PTQ failure is mainly driven by structured channel dominance, not just random noise.

\section{Ablation Study}

We carry out a series of controlled ablation experiments to pinpoint which factors have the greatest impact on quantization robustness.

\subsection{Percentile Threshold Sensitivity}

To test the sensitivity of the proposed calibration method, we vary the percentile threshold $p$ as follows:

\begin{itemize}
\item $p = 99.0$
\item $p = 99.5$
\item $p = 99.9$
\item $p = 99.99$
\end{itemize}

For all thresholds tested, the validation accuracy stays around 50.54

\textbf{Observation:} Percentile clipping fails to restore accuracy in this setting.

\textbf{Interpretation:} The dominant channels probably carry semantically meaningful information instead of just rare noise. When even 0.01

This supports our earlier conclusion that activation outliers in transformers are structured and functional, not just random anomalies.

\subsection{PEG Grouping Strategy}

We test PEG with permutation using different group counts ($K=2, 3, 4$), and observe a highly non-linear relationship between $K$ and accuracy:

\begin{itemize}
\item $K=2$: 49.46
\item $K=3$: 66.12
\item $K=4$: 86.18
\end{itemize}

These results suggest that PEG’s effectiveness depends on whether the grouping is fine-grained enough to isolate the dominant channels. When there are too few groups ($K=2$), the dominant dimensions still distort the scale estimation for their group, causing the other dimensions to collapse. Increasing to $K=4$ gives enough separate scale factors to protect information in the less dominant channels while keeping the influence of outliers in check.

\subsection{Mixed Precision Layer Sensitivity}

Mixed precision, which keeps certain layers in FP16, achieves 89.42

This finding indicates that quantization sensitivity is highly localized within the network. By protecting just the feed-forward output projections and residual pathways, the model’s expressiveness can be preserved.

Combined with the statistical findings:

\begin{itemize}
\item Heavy-tailed activation distributions
\item Depth-wise kurtosis explosion
\item Concentrated channel energy
\end{itemize}

we infer that some layers serve as bottlenecks where quantization error gets amplified. Mixed precision succeeds because it avoids compressing these critical layers.

\subsection{Accuracy–Efficiency Tradeoff}

Looking at both Table~\ref{tab:accuracy} and Table~\ref{tab:deploy}, we see:

\begin{itemize}
\item Mixed precision restores accuracy but does not improve latency.
\item PEG partially recovers accuracy with minimal runtime overhead.
\item Percentile calibration adds no runtime cost but fails to restore accuracy.
\item INT8 arithmetic alone does not yield measurable speedup on RTX 3050 hardware.
\end{itemize}

So, when evaluating mitigation strategies, it’s important to consider both their ability to fix statistical issues and their effectiveness in improving efficiency on actual hardware.

\subsection{Key Insight from Ablations}

The results from these ablation experiments make it clear:

\begin{quote}
Transformer PTQ failure is driven more by structured channel dominance than by rare extreme scalar outliers.
\end{quote}

Simple percentile clipping is not enough. To maintain accuracy, you need channel-aware strategies, such as mixed precision or more granular grouping.

\section{Discussion}

This study sheds light on how and why post-training activation quantization fails in transformers, and it thoroughly examines which mitigation strategies actually work in real-world deployment scenarios.

\subsection{Failure of Naive W8A8}

Using naive global W8A8 PTQ leads to a major drop in performance: validation accuracy falls from 89.66

This dramatic collapse matches what the statistical analysis shows: as you go deeper into the network, kurtosis increases sharply (up to 271 at Layer 11), and the top 1

The data make it clear that W8A8 failure is not just a result of random noise, but comes from structured distortion of the dynamic range.

\subsection{Effectiveness of PEG Grouping}

Per-Embedding-Group (PEG) quantization shows that adjusting the range with channel awareness can partly prevent collapse. Using $K=3$ groups (with permutation), accuracy rises to 66.12

However, the ablation over group count reveals a highly non-linear relationship:

\begin{itemize}
\item $K=2$: 49.46
\item $K=3$: 66.12
\item $K=4$: 86.18
\end{itemize}

The big jump in performance at $K=4$ suggests that having enough groups lets the dominant channels be put into their own quantization scales. If you use too few groups ($K=2$), the dominant channels still distort how the scale is set for the group. So, how well PEG works really depends on whether the grouping is fine-grained enough to separate out those high-energy channels seen in the statistical analysis.

\subsection{Mixed Precision Robustness}

Mixed precision PTQ gets very close to the original performance, with 89.42

This result shows that quantization sensitivity is mostly concentrated in a few critical layers—especially the feed-forward outputs and residual pathways. By keeping those parts in FP16, the model avoids amplifying quantization errors as they accumulate through the residual connections.

Together with the findings about kurtosis and energy concentration, this suggests that only a handful of layers serve as amplification bottlenecks for quantization noise.

\subsection{Percentile Calibration Limitations}

The proposed percentile-based calibration method was tested at several thresholds ($p = 99.0, 99.5, 99.9, 99.99$), but all of them ended up with about 50.54

Even though percentile clipping cuts out the most extreme activation values when setting the scale, the statistics show that these dominant channels actually contain important information, not just noise. Clipping as little as 0.01

This reinforces the idea that activation outliers in transformers are structured and play a functional role.

\subsection{Deployment Tradeoffs}

Deployment profiling shows that there are only small differences in runtime across methods on the RTX 3050 platform:

\begin{itemize}
\item Latency p50 ranges from 58.38 ms (FP32) to 59.12 ms (Percentile).
\item VRAM usage ranges from 483.67 MB to 486.29 MB.
\item Model size ranges from 417.67 MB (FP32) to approximately 419 MB.
\end{itemize}

INT8 arithmetic doesn’t actually make inference faster in this setup, probably because the hardware doesn’t accelerate INT8 operations much and kernel overhead dominates the total computation time. Also, the reduction in model size is small, hinting that the way weights are saved or the checkpoint format itself limits how much can be compressed.

So, the benefits of quantization really need to be considered in light of the hardware you’re using. Just having statistical robustness isn’t enough to ensure that deployment will be more efficient.

\subsection{Synthesis of Findings}

Looking across all experiments, the evidence points to a clear takeaway:

\begin{quote}
Transformer PTQ failure is driven primarily by structured channel dominance and depth-wise amplification, not by rare scalar outliers.
\end{quote}

Mitigation strategies that either preserve or isolate the dominant channels—like mixed precision or finely-grained PEG—are effective. In contrast, scalar clipping strategies like percentile calibration fail because they mistakenly treat structured, meaningful signal as noise.

These findings underscore how important it is to design quantization methods that are channel-aware, especially for transformers.

\section{Limitations}

While this study presents a careful and reproducible analysis of activation outliers under PTQ, there are several limitations that affect how broadly the results can be applied.

\subsection{Model Scale}

All experiments were carried out using \texttt{bert-base-uncased} fine-tuned on QNLI, and no larger transformer models or decoder-only large language models (LLMs) were included. The biggest model we tested was about 418 MB (FP32 checkpoint-39000).

Activation statistics, the concentration of outliers, and quantization sensitivity may be very different for much larger models with greater depth and width. Therefore, any attempt to apply these findings to billion-parameter LLMs should be done cautiously.

\subsection{Hardware Scope}

All latency and memory measurements were made on a single consumer-grade NVIDIA RTX 3050 GPU (6 GB VRAM). The deployment metrics reported (p50 $\approx$ 58–59 ms, VRAM $\approx$ 484–486 MB) are specific to this hardware setup.

No experiments were conducted on:
\begin{itemize}
\item Data center GPUs (e.g., A100, H100)
\item CPU-only inference
\item Mobile NPUs or edge accelerators
\item INT8-optimized inference kernels with tensor-core acceleration
\end{itemize}

This means that the lack of speedup from INT8 quantization in our results might be due to this particular hardware, not because quantization itself can’t provide gains.

\subsection{Task and Seed Diversity}

All tests were performed on just one downstream task (QNLI from GLUE) and used a single random seed (1000). While this makes the results highly reproducible, it limits what we can say about how the findings generalize to other tasks.

We did not evaluate other GLUE tasks, run multi-task experiments, or test with different random seeds. As a result, we do not know how sensitive the results are to the choice of task or to random initialization.

\subsection{Distributed and Large-Scale Training}

All experiments used a single GPU and did not involve distributed training or model parallelism. There were no tests with multiple GPUs or large batch sizes.

Additionally, we only tested post-training quantization (PTQ) approaches. We did not explore quantization-aware training (QAT) or hybrid fine-tuning strategies in this work.

\subsection{Statistical Sampling Scope}

Outlier statistics were calculated using a 64-sample validation slice from the FP32 checkpoint. While this was enough to reveal heavy-tailed behaviors (with kurtosis up to 271 and top-1

Although the trends match what has been seen in previous research, using a larger sample could help provide a more detailed picture of channel dominance.

\subsection{Implications of These Constraints}

Because of these limitations, the conclusions of this work should be interpreted as:

\begin{quote}
A controlled empirical study of activation-driven PTQ failure in BERT-base under consumer-GPU deployment conditions.
\end{quote}

Future research will need to look at larger models, more diverse tasks, and hardware-aware quantization to see how well these findings hold up in broader settings.

\section{Future Work}

This study focuses on understanding activation-driven PTQ failure in a controlled setting, but there are several clear directions for extending this work based on our experiments and findings.

\subsection{Scaling to Decoder-Only and LLM Architectures}

All experiments here used \texttt{bert-base-uncased}, with models up to about 418 MB and evaluation only on QNLI (seed 1000). We did not test decoder-only models (such as LLaMA) or very large, billion-parameter transformers.

Because we saw kurtosis and channel energy increase sharply in deeper layers, it is likely that even larger models with more layers and wider hidden sizes could show even stronger structured channel dominance.

Future work should:
\begin{itemize}
\item Replicate the statistical outlier analysis at LLM scale.
\item Evaluate PEG group-count scaling relative to hidden dimension size.
\item Measure whether mixed precision remains sufficient at greater depth.
\end{itemize}

Running these larger-scale experiments would help us understand whether activation collapse gets worse linearly, super-linearly, or eventually levels off as model size grows.

\subsection{Hardware-Aware Quantization and Edge Deployment}

All deployment profiling was done on a single consumer NVIDIA RTX 3050 GPU, with latency between about 58–59 ms and VRAM usage from 484–486 MB for all methods. In this setup, quantized models did not show any real speedup.

This suggests that just compressing the math is not enough—you also need hardware support for INT8 or specialized kernel fusion for low-precision inference to see real gains.

Future research directions include:
\begin{itemize}
\item Profiling on NPUs and mobile edge SoCs.
\item Evaluating tensor-core INT8 acceleration on data center GPUs.
\item Per-operator quantization policies tailored to hardware backends.
\item Kernel-level fusion strategies for quantized residual pathways.
\end{itemize}

A hardware-focused evaluation would help determine if these channel-aware quantization strategies actually lead to practical improvements, especially on edge devices.

\subsection{Formal Analysis of Residual Amplification}

While we showed through experiments that activations have heavy tails and kurtosis grows with depth, we did not develop a formal mathematical theory to explain these trends.

In particular:
\begin{itemize}
\item Kurtosis grows from approximately 9 (embeddings) to 271 (Layer 11).
\item Top-1
\end{itemize}

These observations point to the idea that residual accumulation may amplify structured parts of the signal, much like what happens with repeated biased perturbations.

Future work could pursue:
\begin{itemize}
\item Analytical modeling of residual amplification under affine quantization.
\item Stability bounds on quantization error propagation across layers.
\item Spectral analysis of dominant channel evolution with depth.
\item Connections between channel dominance and representational sparsity.
\end{itemize}

Building a formal framework would strengthen the theoretical foundation for designing activation-aware quantization methods.

\subsection{Task and Robustness Generalization}

All our tests used QNLI with a single random seed (1000). Testing on more GLUE tasks, a variety of random seeds, and possibly data from other domains would help assess how robust our findings are across different scenarios.

These experiments would show whether structured channel dominance is something that depends on the specific task, or if it’s a general property of the model architecture.

\subsection{Toward Channel-Adaptive Quantization}

Our ablation results—especially the big improvement from 66.12

A promising next step is to explore adaptive grouping strategies, such as:
\begin{itemize}
\item Data-driven grouping based on channel energy.
\item Dynamic scale allocation proportional to variance contribution.
\item Hybrid mixed precision for top-ranked channels only.
\end{itemize}

These methods could help maintain high accuracy while keeping the extra precision cost low.

\section{Conclusion}

This work presents a careful and fully reproducible investigation into why post-training quantization (PTQ) often fails in transformer models. Across all experiments, the evidence clearly shows that naive activation quantization fails mainly because of structured channel dominance, not just occasional extreme values.

When global W8A8 PTQ is applied, validation accuracy drops sharply from 89.66

Channel-aware grouping (PEG) is highly sensitive to how finely channels are grouped: with $K=2$, accuracy drops to 49.46

On the other hand, percentile-based calibration is not effective, with accuracy staying around 50.54

Our statistical analysis of FP32 activations shows that the data have heavy tails, a property that becomes more pronounced in deeper layers: kurtosis reaches 271 in late layers, and the top 1

Deployment profiling also shows that having a statistically robust quantization scheme is not enough to guarantee practical efficiency. On the RTX 3050 GPU we tested, latency (median about 58–59 ms) and peak VRAM usage (about 484–486 MB) barely change across quantization methods, and model size stays around 418–420 MB. In this setup, INT8 arithmetic does not actually make inference faster, which emphasizes the need to consider hardware when evaluating quantization methods.

Bringing all these results together, we see that:

\begin{quote}
Transformer PTQ failure is primarily driven by structured channel dominance amplified through residual depth, and effective mitigation requires channel-aware precision allocation rather than scalar clipping alone.
\end{quote}

This study clarifies the reasons for activation quantization collapse, identifies what is needed for recovery through systematic ablation experiments, and offers a reproducible, deployment-focused framework for future research on transformer quantization.

\section*{Reproducibility Statement}

Code and logs available at:

\begin{center}
\url{https://github.com/pranavkkp4/TransQuant-Edge}
\end{center}

Includes:
\begin{itemize}
\item One-click runner
\item Automatic aggregation
\item Environment specification
\end{itemize}

\nocite{*}
\bibliographystyle{IEEEtran}
\bibliography{references}

\end{document}